\documentclass{article}

\PassOptionsToPackage{numbers, compress}{natbib}

\usepackage[preprint]{neurips_2026}

\usepackage[utf8]{inputenc} 
\usepackage[T1]{fontenc}    
\usepackage{url}            
\usepackage{booktabs}       
\usepackage{amsfonts}       
\usepackage{nicefrac}       
\usepackage{microtype}      
\usepackage{xcolor}         

\usepackage{lipsum}
\usepackage{amsfonts}
\usepackage{graphicx}
\usepackage{epstopdf}
\usepackage{algorithmic}
\ifpdf
  \DeclareGraphicsExtensions{.eps,.pdf,.png,.jpg}
\else
  \DeclareGraphicsExtensions{.eps}
\fi

\usepackage{amsmath}
\usepackage{amssymb}
\usepackage{mathtools}
\usepackage{amsthm}
\newcommand{\ones}{\mathbf 1}

\newcommand{\argmin}{\mathop{\rm argmin}}

\newtheorem{theorem}{Theorem}[section]
\newtheorem{lemma}{Lemma}[section]

\newtheorem{remark}{Remark}[section]

\def\approxcorrect{\checkmark\kern-1.1ex\raisebox{.89ex}{$\times$}}

\usepackage{amsmath,amsfonts,bm}

\def\eqref#1{equation~\ref{#1}}

\def\1{\bm{1}}

\DeclareMathAlphabet{\mathsfit}{\encodingdefault}{\sfdefault}{m}{sl}
\SetMathAlphabet{\mathsfit}{bold}{\encodingdefault}{\sfdefault}{bx}{n}

\def\gB{{\mathcal{B}}}

\def\gP{{\mathcal{P}}}

\def\gX{{\mathcal{X}}}

\def\sI{{\mathbb{I}}}

\def\sR{{\mathbb{R}}}

\usepackage{microtype}
\usepackage{graphicx}
\usepackage{subfigure}
\usepackage{booktabs}
\newcommand{\inprod}[2]{\left\langle #1, #2 \right\rangle}

\usepackage{tcolorbox}
\newtcolorbox{thmbox}{colback=cyan!5,colframe=white,top=2pt,bottom=2pt,left=0pt,right=2pt}
\newtcolorbox{questionbox}{colback=black!5!white,colframe=white}
\usepackage{enumitem}
\usepackage{multirow}

\usepackage{algorithm}

\usepackage{amsopn}
\DeclareMathOperator{\diag}{diag}

\usepackage{hyperref}       

\title{PINS: Proximal Iterations with Sparse Newton and Sinkhorn for Optimal Transport}

\author{%
  Di Wu\thanks{Equal contribution.} \\
  Department of Mathematics\\
  University of Maryland\\
  College Park, MD 20742 \\
  \texttt{dwuwd@umd.edu} \\
  \And
  Ling Liang\footnotemark[1] \\
  Department of Mathematics\\
  University of Tennessee\\
  Knoxville, TN 37916 \\
  \texttt{liang.ling@u.nus.edu} \\
  \AND
  Haizhao Yang\thanks{Corresponding author.} \\
  Department of Mathematics\\
  Department of Computer Science\\
  University of Maryland\\
  College Park, MD 20742 \\
  \texttt{hzyang@umd.edu} \\
}

\begin{document}

\maketitle

\begin{abstract}
Optimal transport (OT) is a widely used tool in machine learning, but computing high-accuracy solutions for large instances remains costly. Entropic regularization and the Sinkhorn algorithm improve scalability; however, when the regularization parameter is small, Sinkhorn convergence slows, and the iterates approach an entropic solution that remains separated from the true OT plan by an entropic-bias plateau. We introduce \textbf{PINS} (\textbf{P}roximal \textbf{I}terations with sparse \textbf{N}ewton and \textbf{S}inkhorn), a two-loop solver designed to move beyond this plateau. The outer loop applies an entropic proximal-point method, solving the original OT problem through a sequence of entropic subproblems with shifted cost matrices. Each inner subproblem is then solved by a Sinkhorn warm-up followed by sparse-Newton refinement. We prove that PINS converges globally to an optimal solution of the unregularized OT problem and that the inner Hessian admits a sparsification at every outer iteration with a structure independent of the cost matrix. On synthetic and augmented-MNIST instances, PINS achieves much lower relative cost errors than Sinkhorn-type baselines, which stall at the entropic-bias plateau, and is $5$--$73\times$ faster than Sinkhorn with the same outer loop at matched accuracy. On large-scale DOTmark instances, a streaming implementation reduces peak memory by $24$--$54\%$ compared with the network-simplex linear programming (LP) solver and remains feasible under per-process memory budgets for which the LP solver fails.
\end{abstract}

\section{Introduction}
\label{intro}
Optimal transport (OT) \citep{villani2009optimal,peyre2019computational} provides a mathematical framework for comparing probability distributions and has found wide use in economics \citep{galichon2018optimal}, physics \citep{de2015power}, imaging \citep{papadakis2015optimal}, and machine learning \citep{peyre2019computational}. At its core, OT seeks the least-cost way to transport mass from one probability distribution to another. In machine learning, OT has been used for generative modeling, domain adaptation, representation learning, and related tasks, which has led to sustained interest in scalable algorithms for large-scale OT problems \citep{sandler2011nonnegative,jitkrittum2016interpretable,arjovsky2017wasserstein,salimans2018improving,genevay2018learning,gao2019deep,chen2020graph,wang2021deep,fatras2021unbalanced,yang2024corrected,hou2024sparse,liang2024accelerating,zhu2024ripalm}. Despite this progress, computing \emph{accurate} solutions for large instances remains a major bottleneck.

We consider the classical discrete Kantorovich OT problem~\citep{villani2009optimal}:
\begin{equation}
    \label{ot}
    \min_{X\in\sR^{m\times n}}\;\inprod{C}{X}\;\;
    \mathrm{s.t.}\;\;Xe_n=a,\;X^{\top}e_m=b,\;X\geq 0,
\end{equation}
where $C\in\sR^{m\times n}$ is the cost matrix, $a\in\sR^m_{++}$, $b\in\mathbb{R}^n_{++}$, and $e_m^{\top}a=e_n^{\top}b=1$. When the two discrete distributions each have $n$ atoms, the decision variable has dimension $\mathcal{O}(n^2)$. Standard algorithms, such as the simplex method \citep{burkard2012assignment} and interior-point methods \citep{nesterov1994interior}, can incur per-iteration computational costs on the order of $\mathcal{O}(n^6)$, making them impractical for large-scale instances.

Entropic regularization with parameter $\eta>0$, also known as Schr\"odinger's problem~\citep{leonard2013survey}, makes OT amenable to Sinkhorn iteration~\citep{cuturi2013sinkhorn} and is now the dominant scalable approach. However, two limitations impose a hard ceiling on the accuracy and parameter regimes that it can reliably reach: \\
(i) \textit{Entropic-bias plateau.} Sinkhorn returns the optimizer of the entropically regularized problem, whose objective value differs from that of the unregularized OT problem by $\mathcal{O}(\eta\log n)$. \\
(ii) \textit{Slow contraction at small $\eta$.} The Sinkhorn
iteration's contraction rate in Hilbert's projective metric tends to one as $\eta\to 0$, and its iteration count to a fixed marginal tolerance scales as $\Theta(1/\eta)$~\citep{altschuler2017near,lin2022efficiency}. Log-domain implementations~\citep{schmitzer2019stabilized} improve numerical stability but do not change this rate.

Together, these effects make the regime of efficient, high-accuracy computation difficult to access. This motivates the following question:\\
\textit{How can one solve the original unregularized OT problem to high accuracy, efficiently, and with reduced sensitivity to $\eta$?}

We answer this question with \textbf{PINS}: \textbf{P}roximal \textbf{I}terations with sparse \textbf{N}ewton and \textbf{S}inkhorn. PINS embeds entropic regularization within a Bregman entropic proximal-point outer loop (EPPA). The main idea is to solve the original OT problem through a sequence of entropy-regularized OT subproblems with shifted cost matrices. Each subproblem is solved in two phases: a Sinkhorn warm-up brings the dual variables into the attraction region of Newton's method, and a sparse-Newton refinement then achieves fast local convergence using a sparsified Hessian. Our contributions are as follows:

\textbf{Algorithm.} We propose PINS, a proximal scheme that couples Sinkhorn warm-up with a sparsified Newton inner solver inside an EPPA outer loop. The outer loop separates the target problem, namely the original unregularized OT problem, from the regularization used in the inner solver, namely entropy with fixed $\eta$. As a result, the choice of $\eta$ no longer determines the final accuracy. PINS computes highly accurate solutions of the original OT while retaining strong efficiency and stability.

\textbf{Theory.} We prove two main results. First, PINS converges globally to an optimal solution of~(\ref{ot}) under summable inexactness. Second, at every outer iteration, the inner Hessian admits a cost-independent sparsification. This structural result makes the per-iteration savings accumulate.

\textbf{Experiments.} On synthetic and augmented-MNIST instances (Section \ref{sec:numerical_1}), PINS reaches relative errors of $10^{-4}$--$10^{-5}$, while Sinkhorn-type baselines stall at the $\mathcal{O}(\eta)$ plateau around $10^{-1}$--$10^{-2}$. At matched accuracy, PINS is $5$--$73\times$ faster than Sinkhorn-with-EPPA. An $\eta$-sweep at $n=400$ further shows that PINS accurately solves the linear programming (LP) at $\eta\in\{10^{-3},10^{-4}\}$, whereas Sinkhorn exhausts its $50{,}000$-iteration cap. On DOTmark instances up to $n=65{,}536$ (Section \ref{sec:numerical_3}), PINS reduces peak resident memory by $24$--$54\%$ compared with the network-simplex LP solver, and there are per-process memory budgets under which PINS is the only feasible solver.

\section{Related Works}
\textbf{Computational OT} Since the introduction of the Sinkhorn algorithm~\citep{cuturi2013sinkhorn}, many methods have been developed to improve the scalability of OT solvers. Representative advances include Newton-accelerated methods~\citep{tangaccelerating}, constrained OT solvers~\citep{tang2024sinkhorn}, and robust sparsification techniques~\citep{tang2024safe}. Complementary approaches improve tractability by accepting approximate solutions, including stochastic optimization~\citep{altschuler2017near}, low-rank factorization~\citep{scetbon2021low}, kernel approximation~\citep{solomon2015convolutional,altschuler2019massively,scetbon2020linear,huguet2023geodesic}, sliced-Wasserstein projections~\citep{bonneel2015sliced}, and alternative entropy penalties~\citep{benamou2015iterative}. These methods target the entropic problem itself or a relaxation of it, so their final
accuracy is limited by the entropic bias. PINS differs from this line of work by directly solving the original unregularized OT problem.

\textbf{OT in machine learning.}
OT-based losses are widely used in generative modeling and distributional alignment, including Wasserstein GANs~\citep{arjovsky2017wasserstein,salimans2018improving}, generative learning with regularized OT~\citep{genevay2018learning,gao2019deep}, OT-aware diffusion and flow-matching models~\citep{de2021diffusion,wang2021deep,shi2024diffusion}, multimodal alignment~\citep{wang2023large}, domain adaptation~\citep{fatras2021unbalanced,yang2024corrected}, graph representation learning~\citep{chen2020graph,hou2024sparse}, and distributional reinforcement learning~\citep{bellemare2017distributional,dabney2018distributional,luo2023optimal,wu2024neural,kulinski2023towards}. We refer readers to~\citet{torres2021survey,montesuma2024recent,peyre2019computational} for broader surveys. In many of these applications, the OT solver is embedded within a larger learning pipeline, where solver error can propagate to downstream objectives. This provides a strong motivation for OT solvers that can achieve accuracy below the entropic-bias plateau.

\textbf{Bregman proximal methods.}
Bregman-distance proximal-point algorithms have a long history in linearly constrained optimization~\citep{censor1992proximal,chen1993convergence,eckstein1993nonlinear,eckstein1998approximate} and have recently been used to design fast linear-programming solvers~\citep{xie2020fast,yang2022bregman,chu2023efficient,yang2025inexact}. Our algorithm is an entropic proximal method on the OT polytope equipped with an inner solver and the cost-independent sparsification at each outer step.

\section{Methodology} \label{sec:method}

\subsection{The Sinkhorn Algorithm}
 \label{sinkhorn}
 The Sinkhorn algorithm solves the entropy-regularized OT
 problem~\citep{cuturi2013sinkhorn}
 \begin{align}
     \label{eot}
     \min_{X\in\sR^{m\times n}_+}\; \inprod{C}{X}
     + \eta \sum_{i,j}X_{ij}\log X_{ij}
     \quad\mathrm{s.t.}\quad Xe_n = a,\; X^\top e_m = b,
 \end{align}
 with $\eta>0$. Let $E\in\sR^{m\times n}$ be the all-ones matrix. Lagrangian duality implies primal recovery map:
 \begin{align}
\label{eq:sink_x}
    X(f,g) = \exp\left(\frac{1}{\eta}\left(fe_n^{\top} + e_mg^{\top} - C - \eta E\right)\right),
\end{align}
where $f\in \sR^m$ and $g\in \sR^n$ are dual variables. The resulting smooth concave dual objective $P(f,g)$ is maximized by block-coordinate ascent. Its log-domain implementation~\citep{schmitzer2019stabilized}, shown in Algorithm~\ref{alg-sinkhorn}, controls floating-point overflow and is the standard form used in modern OT pipelines.

\begin{algorithm}[!h]
    \begin{algorithmic}[1]
        \STATE \textbf{Inputs:} cost $C\in\sR^{m\times n}$, marginals
        $(a,b)\in\sR^{m+n}_{++}$, dual variables $(f^0,g^0)\in\sR^{m+n}$, $\eta>0$.
        \FOR{$k\geq 0$}
            \STATE \(
                 f^{k+1} \leftarrow  f^k + \eta\bigl(\log a
                 - \log\bigl(X(f^k, g^k)\,e_n\bigr)\bigr).
            \)
            \STATE \(
                 g^{k+1} \leftarrow g^k + \eta\bigl(\log b
                 - \log\bigl(X(f^{k+1}, g^k)^{\top}e_m\bigr)\bigr).
            \)
        \ENDFOR
        \STATE \textbf{Output:} $X(f^{k+1},\,g^{k+1})$.
    \end{algorithmic}
    \caption{Log-domain Sinkhorn.}
    \label{alg-sinkhorn}
\end{algorithm}

\textbf{Two structural limitations.}
Algorithm~\ref{alg-sinkhorn} solves the entropic
problem~(\ref{eot}), not the original LP~(\ref{ot}). Two limitations restrict the accuracy attainable within a fixed wall-clock budget. First, \emph{entropic-bias plateau}: the optimum of the entropic problem differs from the LP optimum by $\mathcal{O}(\eta\log n)$, so reducing the inner tolerance below this accuracy floor does not improve the solution to the original OT problem. Second, \emph{slow contraction at small $\eta$}: the contraction rate of Sinkhorn iteration in Hilbert's projective metric tends to one as $\eta\to 0$, and the number of inner iterations required to reach a fixed marginal tolerance scales as $\Theta(1/\eta)$~\citep{altschuler2017near,lin2022efficiency}. Together, these effects make the regime of small $\eta$ and high accuracy difficult to reach for Sinkhorn-type methods, as we verify empirically in Section \ref{sec:numerical_1_eta}.

\subsection{PINS } \label{sec:pins}
PINS separates two computational roles: an \emph{outer loop} that solves the original LP~(\ref{ot}) by reducing it to a sequence of entropic subproblems with shifted cost matrices, and an \emph{inner solver} that combines the global stability of Sinkhorn with the fast local convergence of Newton's method for each subproblem.

\begin{algorithm}[!h]
    \begin{algorithmic}[1]
        \STATE \textbf{Inputs:} cost $C\in\sR^{m\times n}$, marginals
        $a,b$, dual variables $(f^0,g^0)$, primal points
        $X^0\in\sR^{m\times n}_{++}$, $\eta>0$, sparsity
        threshold $\rho>0$, iteration caps $T_1$ (Sinkhorn), $T_2$
        (Newton), $K$ (EPPA outer).
        \FOR{$k = 0,\ldots,K-1$}
            \STATE \(C^k \leftarrow C - \eta \log X^k\)
            \hfill // shifted cost for the $k$-th subproblem
            \STATE \(\textit{$(f,g) \leftarrow (f^0,g^0)$ if $k=0$,
                else warm-start from previous outer iterate.}\)
            \FOR{$t = 0,\ldots,T_1-1$}
                \STATE \(
                    f \leftarrow f + \eta\bigl(
                        \log a - \log\bigl(\widetilde{X}^k(f,g)\,e_n\bigr)
                    \bigr)\)
                \STATE \(
                    g \leftarrow g + \eta\bigl(
                        \log b - \log\bigl(\widetilde{X}^k(f,g)^{\top}e_m\bigr)
                    \bigr)\)
            \ENDFOR
            \FOR{$t = 0,\ldots,T_2-1$}
                \STATE \(H \leftarrow \mathrm{Sparsify}\bigl(
                    \nabla^2 P^k(f,g),\,\rho\bigr)\)
                \STATE \(
                    (\Delta f,\Delta g) \leftarrow
                    \mathrm{CG}\bigl(H,\,-\nabla P^k(f,g)\bigr)\)
                \STATE \(
                    \alpha \leftarrow \mathrm{LineSearch}\bigl(
                        P^k,\,(f,g),\,(\Delta f,\Delta g)\bigr)\)
                \STATE \((f,g) \leftarrow (f,g) + \alpha\,(\Delta f,\Delta g)\)
            \ENDFOR
            \STATE \(X^{k+1} \leftarrow \widetilde{X}^k(f,g)\)
        \ENDFOR
        \STATE \textbf{Output:} $X^{K}$.
    \end{algorithmic}
    \caption{PINS: Proximal Iterations with sparse Newton and Sinkhorn.}
    \label{alg-PINS}
\end{algorithm}

\textbf{Outer loop.}
The Boltzmann--Shannon entropy
\(
    \phi(X):=\sum_{i,j}X_{ij}\bigl(\log X_{ij}-1\bigr),
\)
induces the Bregman divergence
\(
    D_\phi(X,Y) := \phi(X)-\phi(Y)-\inprod{\nabla\phi(Y)}{X-Y},
\)
which is non-negative on $\sR^{m\times n}_{+}$ and vanishes iff
$X=Y$. Applying the entropic proximal-point algorithm~\citep{xie2020fast,chu2023efficient} to~(\ref{ot}) gives
\begin{align}
    \label{eq:x_inner}
    X^{k+1}\;\in\;\argmin\Bigl\{\,
        \inprod{C}{X}+\eta D_\phi(X,X^k)
        \;:\; Xe_n=a,\; X^{\top}e_m=b
    \,\Bigr\}.
\end{align}
Theorem~\ref{thm:convergence} shows that under summable inexactness, $\{X^k\}$ converges to an optimum of the original problem~(\ref{ot}), instead of the solution of a fixed entropic surrogate. Since every feasible $X$ in~(\ref{ot}) satisfies $\sum_{ij}X_{ij}=1$, expanding $D_\phi$ and dropping constants reduces~(\ref{eq:x_inner}) to the entropic OT subproblem 

\begin{align}
    \label{eppa-sub}
    \min_{X\in\sR^{m\times n}_+}\;
        \inprod{C^k}{X}+\eta\sum_{i,j}X_{ij}\log X_{ij}
    \quad\mathrm{s.t.}\quad Xe_n=a,\; X^\top e_m=b,
\end{align}
where $C^k:=C-\eta\log X^k$ is the shifted cost. Thus, each outer iteration requires solving an entropic OT instance with the same structure as~(\ref{eot}) but with a cost matrix that changes with $k$.

\textbf{Inner solver.}
We solve~(\ref{eppa-sub}) through its dual. Setting
$\nabla_X L^k = 0$ for the Lagrangian
\(
    L^k(X,f,g)
    = \inprod{C^k}{X}+\eta\sum_{ij}X_{ij}\log X_{ij}
    + \inprod{f}{a-Xe_n}+\inprod{g}{b-X^\top e_m}
\)
gives the primal recovery map
\begin{align}
    \label{eq:eppa_x}
    \widetilde{X}^k(f,g)
    = \exp\!\left(\tfrac{1}{\eta}\bigl(
        fe_n^{\top}+e_mg^{\top}-C^k-\eta E\bigr)\right),
\end{align}
and the smooth and strictly concave dual problem is
\begin{align}
    \label{eppa-sub-dual}
    \max_{f\in\sR^m,\,g\in\sR^n}\;
    P^k(f,g) = \inprod{a}{f} + \inprod{b}{g}
    - \eta\,\inprod{E}{\widetilde{X}^k(f,g)}.
\end{align}
Direct calculation gives
\(
    \nabla_f P^k = a - \widetilde{X}^k(f,g)\,e_n,
\)
\(
    \nabla_g P^k = b - \widetilde{X}^k(f,g)^{\top}e_m,
\)
and the Hessian
\begin{align}
    \label{eq:eppa-hess}
    \nabla^2 P^k(f,g)
    = -\frac{1}{\eta}\begin{bmatrix}
        \diag\!\bigl(\widetilde{X}^k(f,g)\,e_n\bigr) &
            \widetilde{X}^k(f,g) \\[2pt]
        \widetilde{X}^k(f,g)^{\top} &
            \diag\!\bigl(\widetilde{X}^k(f,g)^{\top}e_m\bigr)
    \end{bmatrix}.
\end{align}
PINS solves~(\ref{eppa-sub-dual}) in two phases: $T_1$ Sinkhorn
block-coordinate ascents bring $(f,g)$ into Newton's region of
attraction, then $T_2$ damped Newton steps on a sparsified Hessian
converge quadratically. The next outer iterate is recovered as
$X^{k+1}=\widetilde{X}^k(f^{T_1+T_2},\,g^{T_1+T_2})$. Across outer
iterations, the inner-loop initial point $(f^0,g^0)$ is warm-started
from the previous outer iterate's terminal duals.

\textbf{Implementation details.}
The operator $\mathrm{Sparsify}(\cdot,\rho)$ retains only entries above a relative threshold $\rho$ in the row--column coupling blocks of $\nabla^2 P^k$; Theorem~\ref{thm:sparse} shows this sparsification is structurally justified. The resulting symmetric sparse system is solved by truncated conjugate gradient~\citep{fletcher1964function,nocedal1999numerical}, and step sizes are chosen by Armijo backtracking line search. The streaming implementation used in the large-scale experiments, such as those in Section~\ref{sec:numerical_3}, computes Hessian-vector products through chunked log-sum-exp and never materializes the dense cost matrix or the dense Hessian.

\section{Theoretical Guarantee}
In this section, we establish the global convergence of PINS and analyze the Hessian sparsification at each outer iteration. These results explain two key aspects of the empirical performance of PINS. Theorem~\ref{thm:convergence} shows that the outer loop drives $\{X^k\}$ to an optimum of the original problem~(\ref{ot}), independently of the choice of $\eta$. Theorem~\ref{thm:sparse} shows that the inner Hessian admits a cost-independent sparsification at each outer iteration. Together, these results show that the outer loop removes the entropic bias, while the inner sparsification keeps each outer step computationally efficient.

\textbf{Notation.}
For a matrix $A$, let $\tau(A)$ denote the fraction of its nonzero entries. Let $\mathcal{P}:=\{X\in\sR^{m\times n}_{+}:Xe_n=a,\,X^\top e_m=b\}$ denote the transport polytope, and let $\Delta>0$ be the cost gap between an optimal vertex of $\mathcal{P}$ and the next-best vertex with respect to the reference cost $C$. The optimal vertex is unique almost surely under continuous cost distributions~\citep{dieci2019boundary,bertsimas1997introduction}; we assume this uniqueness throughout. Additional notation used only in the proofs is introduced in the appendix.

\subsection{Global convergence of the outer loop}
\label{sec:thm1}
Each subproblem~(\ref{eq:x_inner}) is solved approximately by the inner procedure described in Section~\ref{sec:pins}. We model this inexactness at outer iteration $k+1$ through the perturbed first-order optimality condition
\begin{align}
\label{eq:inexactness}
    \delta^{k+1}\;\in\;
    \partial \sI_{\Omega}(X^{k+1})
    + C
    + \eta\bigl(\nabla\phi(X^{k+1})-\nabla\phi(X^{k})\bigr),
\end{align}
where $\sI_{\Omega}$ is the indicator function of the feasible set $\Omega=\{X:Xe_n=a,\,X^\top e_m=b\}$, $\partial f$ denotes the subdifferential of $f$, and the residuals satisfy $0\leq \|\delta^{k+1}\|_F \leq \epsilon^k, \sum_{k\geq 0}\epsilon^k<\infty.$
This condition can be interpreted as an $\epsilon$-stationary version of the exact first-order optimality condition.

\begin{theorem}
\label{thm:convergence}
Let $\{X^k\}$ be generated by~(\ref{eq:x_inner}) with inexactness
modeled by~(\ref{eq:inexactness}). Then $\{X^k\}$ converges to an
optimal solution of the original unregularized OT problem~(\ref{ot}).
\end{theorem}

\begin{proof}[Proof sketch]
Setting $P=X^k$ in~(\ref{eq:inexactness}) with Bregman
three-point identity yields the descent inequality
$\inprod{C}{X^{k+1}}\leq\inprod{C}{X^k}
+ \sqrt{2\max\{m,n\}}\,\epsilon^{k+1}$. Summability of
$\{\epsilon^k\}$ then gives convergence of $\{\inprod{C}{X^k}\}$.
Setting $P=X^*$ for any optimum $X^*$ yields a quasi-Fej\'er bound
on $D_\phi(X^*,X^k)$; a subsequence argument together with Pinsker's
inequality promotes weak limit-point optimality to convergence of the
full sequence in $\ell_1$ to an optimum. The full proof is in
Appendix~\ref{app:proof_thm1}.
\end{proof}

Three properties of this result drive the algorithm design. First, the limit is an optimum of the unregularized problem, not of an $\eta$-regularized surrogate. Thus, PINS is not constrained by the entropic-bias plateau that limits Sinkhorn. Second, the result holds for any fixed $\eta>0$, so $\eta$ can be chosen according to the needs of the inner solver, such as the Newton region and sparsification quality, rather than as a final-accuracy parameter. Third, the summable inexactness condition $\sum_k \epsilon^k<\infty$ allows finite-precision inner solves with controlled cost. The combination allows PINS to attain accuracy for the original OT while using a cheap inner solver at any positive regularization level.

\subsection{Cost-independent inner sparsification}
\label{sec:thm2}
We now analyze the inner subproblem. Recall that the dual Hessian~(\ref{eq:eppa-hess}) at an iterate $\widetilde{X}^k$ is generally dense, with $\mathcal{O}((m+n)^2)$ off-diagonal entries. The next theorem shows that a sparse approximation structure exists at \emph{every} outer step independent of $k$ and of $C$, and bounds its $\ell_1$ approximation error.

\begin{theorem}
\label{thm:sparse}
Consider outer iteration $k$ of PINS, and let $\widetilde{X}^k$ denote
the Sinkhorn warm-up output of the inner procedure. Let
$X_{\gX}^{k*}$ be the closest optimum of the $k$-th subproblem
to $\widetilde{X}^k$ in $\ell_1$. Then there exist constants
$\kappa,\,l$, depending only on the subproblem, such that whenever
the Sinkhorn warm-up performs $T_1>\kappa$ steps and the regularization
satisfies $\eta\leq\Delta/(1+\log(mn))$, one can construct a sparse
matrix $H^k\in\sR^{(m+n)\times(m+n)}$ with
\begin{align*}
    \tau(H^k)\;\leq\;
        \frac{2mn\,\tau(X_{\gX}^{k*})+m+n}{(m+n)^2}
    \quad\text{and}\quad
    \bigl\|H^k - \nabla^2 P^k(\widetilde{X}^k)\bigr\|_1
    \;\leq\;\eta^{-1}\!\left(6mn\,e^{-\Delta/\eta}
                              + \sqrt{l/T_1}\right).
\end{align*}
In particular, when $m=n$,
\(
    \tau(H^k)\leq (3n-1)/(2n^2)
\)
almost surely, independent of $k$ and $C$.
\end{theorem}

\begin{proof}[Proof sketch]
Construct
$H^k=-\eta^{-1}\bigl[\diag(\widetilde{X}^k e_n),\,
X_{\gX}^{k*};\,X_{\gX}^{k*\top},\,\diag(\widetilde{X}^{k\top}e_m)\bigr]$,
which has the claimed sparsity by counting nonzeros. The
approximation error reduces to
$\eta^{-1}\|\widetilde{X}^k-X_{\gX}^{k*}\|_1$, which is bounded by the
triangle inequality through a Sinkhorn warm-up error
$\sqrt{l/T_1}$~\citep{ghosal2022convergence} and a Weed-style
deviation $d_1(\widetilde{X}^{k*},\gX^k)\leq 6mn\,e^{-\Delta/\eta}$. The $m=n$ density bound follows since
optima are vertices of the Birkhoff polytope and have at most
$2n-1$ nonzeros~\citep{stanley2011enumerative,bertsimas1997introduction}.
Full proof in Appendix~\ref{app:proof_thm2}.
\end{proof}

The sparsity bound holds for all outer iterations and is independent of the cost matrix. Hence, the per-iteration cost saving accumulates across the EPPA outer loop, in contrast with sparsification schemes whose density bounds degrade away from a single warm point. This behavior is empirically reflected in the size-monotone speedups reported later: as $n$ increases, both the cost of each sparse-Newton step and the number of outer iterations needed to reach the target accuracy remain favorable.

The error bound in Theorem~\ref{thm:sparse} has an $\eta^{-1}$ prefactor and an $e^{-\Delta/\eta}$ term. As $\eta\to 0$, the exponential dominates, so smaller $\eta$ improves sparsification accuracy. As $\eta$ approaches $\Delta/(1+\log(mn))$, the admissible range becomes restrictive. Outside PINS, this would require careful tuning of $\eta$ against the target accuracy, since Sinkhorn incurs an entropic bias for any fixed choice of $\eta$. In PINS, however, the EPPA outer loop decouples $\eta$ from the final accuracy by Theorem~\ref{thm:convergence}. Thus, $\eta$ can be selected according to sparsification quality and the Newton attraction region alone. This explains the later empirical observation: PINS remains accurate across different orders of magnitude in $\eta$, whereas Sinkhorn either reaches the entropic-bias plateau for large $\eta$ or exhausts its iteration cap for small $\eta$.

\section{Numerical Experiments}
We evaluate PINS on synthetic and augmented-MNIST instances against
four Sinkhorn-/Newton-type baselines (\S\ref{sec:numerical_1}) and on
large-scale DOTmark benchmarks with a memory-feasibility study against
the network-simplex LP solver (\S\ref{sec:numerical_3}). A qualitative color-transfer example is provided in Appendix~\ref{app:color_transfer}.

\subsection{Comparison on Synthetic and MNIST Datasets} \label{sec:numerical_1}

\subsubsection{Datasets and Methods}\label{sec:numerical_1_setup}
\textit{Synthetic.} The random assignment problem with additional constraints is widely studied in the literature \cite{mezard1987solution,steele1997probability,aldous2001zeta}. Specifically, we generate an $n\times n$ cost matrix $C$ with entries $c_{ij}\sim \text{Unif}([0,1])$. The source and target vectors are defined as $a=b=\frac{1}{n}\ones$. The values of $n$ are set to $\{50,100,200,400\}$.

\textit{Augmented MNIST.} For each grid size $N\in\{1,2,4,8\}$, we tile $N^2$ MNIST images \citep{lecun1998gradient} into a single $28N\times 28N$ grayscale image. Sampling two tiled images defines the source and target distributions $a$ and $b$. Background-zero pixels are removed, and the remaining intensities are renormalized to sum to one. This setup creates two main challenges. First, the cost matrix becomes large. Second, the tiled images contain gaps between sub-images, which induce sparse graph structures and disconnected regions that restrict mass transport paths, making the problem harder to solve~\citep{cipolla2024regularized}.

\textbf{Methods.} Four methods are compared: (1) Sinkhorn (no EPPA). Log-domain Sinkhorn at fixed $\eta$. (2) Newton (no EPPA). The sparse Newton inner solver followed by truncated CG on the sparsified Hessian. (3) Sinkhorn + EPPA. The same EPPA outer loop as PINS, with Sinkhorn as the inner solver. (4) PINS. EPPA outer loop with the sparse Newton inner solver at every outer step.

\subsubsection{Exact optimality}
\label{sec:numerical_1_accuracy}
\begin{figure}[!h]
\centering
\includegraphics[width=\textwidth]{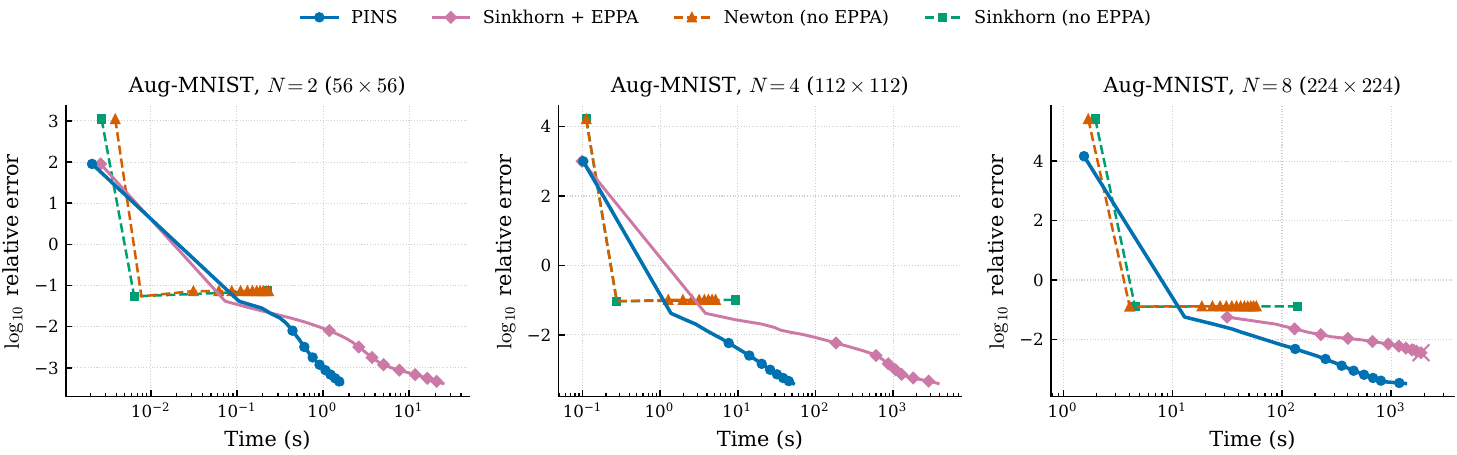}
\caption{$\log_{10}$ relative error to the optimum on augmented MNIST ($\eta = 10^{-1}$). The x-axis is $\log_{10}$ wall-clock time (seconds). The figure shows two main findings. First, the two no-EPPA baselines quickly descend to the entropic-bias plateau ($\log_{10}$ gap $\approx -1$) and stall, because the inner solver cannot remove the bias of the $\eta$-regularized problem. PINS and Sinkhorn + EPPA, sharing the EPPA outer loop, both move past this plateau toward a $\log_{10}$ gap $\approx -3.4$. Second, at $N=8$, the ``$\times$'' marker on the Sinkhorn + EPPA curve indicates that the method reaches the $1{,}800$\,s wall-clock cap before matching the final accuracy of PINS: PINS reaches a $\log_{10}$ gap of $-3.48$ in $1{,}344$\,s. Together with the first two panels, this shows that PINS is both faster and more accurate at this size.}
\label{fig:mnist_combined}
\end{figure}

Figure~\ref{fig:mnist_combined} shows the per-iteration $\log_{10}$ relative error of the transport cost versus cumulative wall-clock time on augmented MNIST for $N\in\{2,4,8\}$. PINS reaches $\log_{10}$ gaps between $-4.5$ and $-4.7$ on synthetic instances and between $-3.3$ and $-3.5$ on MNIST. In contrast, both Newton without EPPA and Sinkhorn without EPPA stall in the range $-1$ to $-2.5$. Numerical values across all problem sizes are summarized in Table~\ref{tab:final_accuracy}. The synthetic per-iteration trajectories are reported in Appendix~\ref{app:section_5_1_extra_figures}.

\begin{remark}
    The initial points shown for the two methods in the figures are not identical. This difference arises because PINS initializes the Lagrange multipliers $f,g$ as zero vectors, while the plotted error is the difference in computed transport cost. Since the cost matrices $C$ and $C^k$ differ, the solutions recovered from~(\ref{eq:sink_x}) and~(\ref{eq:eppa_x}) also differ. This leads to the apparent mismatch in initialization.
\end{remark}



\begin{table}[!ht]
\caption{Final relative error to the optimum at the default
$\eta$ for each family ($\eta = 10^{-2}$ on synthetic,
$\eta = 10^{-1}$ on MNIST), reported as $\log_{10}$ relative
error of the transport cost. The Newton/Sinkhorn-without-EPPA columns share the same plateau because both stop at the entropic-bias accuracy set by $\eta$, regardless of the inner solver. PINS' final accuracy reflects the configured outer tolerance.}
\label{tab:final_accuracy}
\centering
\small
\begin{tabular}{lrrrr}
\toprule
Dataset & PINS & Newton (no EPPA) & Sinkhorn (no EPPA) & Gap factor \\
\midrule
Synthetic, $n = 50$    & $\mathbf{-4.54}$ & $-2.51$ & $-2.51$ & $\sim 100\times$ \\
Synthetic, $n = 100$   & $\mathbf{-4.65}$ & $-2.37$ & $-2.37$ & $\sim 200\times$ \\
Synthetic, $n = 200$   & $\mathbf{-4.67}$ & $-2.22$ & $-2.22$ & $\sim 280\times$ \\
Synthetic, $n = 400$   & $\mathbf{-4.75}$ & $-2.12$ & $-2.12$ & $\sim 430\times$ \\
MNIST, $N = 1$         & $\mathbf{-3.31}$ & $-1.17$ & $-1.17$ & $\sim 140\times$ \\
MNIST, $N = 2$         & $\mathbf{-3.38}$ & $-1.13$ & $-1.13$ & $\sim 180\times$ \\
MNIST, $N = 4$         & $\mathbf{-3.40}$ & $-0.99$ & $-0.99$ & $\sim 260\times$ \\
MNIST, $N = 8$         & $\mathbf{-3.48}$ & $-0.89$ & $-0.89$ & $\sim 390\times$ \\
\bottomrule
\end{tabular}
\end{table}

\subsubsection{Time efficiency}
\label{sec:numerical_1_time}
Table~\ref{tab:speedups} reports the wall-clock time of PINS and Sinkhorn + EPPA at matched final accuracy, using the same EPPA outer iteration count. PINS is $5$--$33\times$ faster on synthetic and $5$--$73\times$ faster on MNIST. The speed-up increases monotonically with problem size on both data families. The PINS and Sinkhorn + EPPA per-iteration trajectories on MNIST are overlaid with the no-EPPA baselines in Figure~\ref{fig:mnist_combined}. The corresponding synthetic results are provided in Appendix~\ref{app:section_5_1_extra_figures}.

\begin{table}[!h]
\caption{Wall time and matched-accuracy speed-up of PINS over
Sinkhorn + EPPA. Speed-up is the ratio of Sinkhorn + EPPA wall
to PINS wall at the same final $\log_{10}$ relative error to
within $10^{-3}$.}
\label{tab:speedups}
\centering
\small
\begin{tabular}{lrrr}
\toprule
Dataset & PINS (s) & Sinkhorn + EPPA (s) & Speed-up \\
\midrule
Synthetic, $n = 50$               & $\mathbf{0.086}$   & $0.485$                & $5.7\times$  \\
Synthetic, $n = 100$              & $\mathbf{0.181}$   & $2.20$                 & $12.2\times$ \\
Synthetic, $n = 200$              & $\mathbf{0.564}$   & $10.88$                & $19.3\times$ \\
Synthetic, $n = 400$              & $\mathbf{1.455}$   & $47.68$                & $32.8\times$ \\
MNIST, $N = 1$                    & $\mathbf{0.259}$   & $1.17$                 & $4.5\times$  \\
MNIST, $N = 2$                    & $\mathbf{1.66}$    & $24.66$                & $14.8\times$ \\
MNIST, $N = 4$                    & $\mathbf{51.74}$   & $3{,}793.2$            & $73.3\times$ \\
\bottomrule
\end{tabular}
\end{table}
These results are consistent with Theorem~\ref{thm:sparse}. The approximate Hessian has bounded sparsity independently of the cost matrix and at every outer iteration. This allows the sparse Hessian system to be solved efficiently by conjugate gradient. In contrast, Sinkhorn + EPPA solves each subproblem using a first-order method, which leads to higher computational cost reflected in the timing results.

\subsubsection{Regularization parameter}
\label{sec:numerical_1_eta}
Table~\ref{tab:eta_sweep} compares PINS and Sinkhorn without EPPA over $\eta\in\{10^{-1},10^{-2},10^{-3},10^{-4}\}$ on synthetic instances with $n=400$. As $\eta$ decreases, the final gap of PINS improves monotonically from $-4.18$ to $-6.13$. Its wall-clock time is nonmonotone because smaller $\eta$ reduces the number of outer iterations, from $242$ to $4$, but increases the cost of each inner Newton solve. Sinkhorn faces two problems at small $\eta$. First, its final gap is bounded below by the entropic-bias plateau, so it cannot match the accuracy of PINS at any tested value of $\eta$. Second, its iteration count to reach a fixed marginal tolerance scales as $\Theta(1/\eta)$~\citep{altschuler2017near}. At $\eta=10^{-3}$ and $10^{-4}$, the contraction is slow enough that Sinkhorn reaches the $50{,}000$-iteration cap after $267$\,s and $214$\,s, respectively, without satisfying the marginal tolerance. The advantage of PINS is therefore largest in the small-$\eta$ regime: at $\eta=10^{-3}$, PINS is approximately $78\times$ faster and $1.8$ decades more accurate; at $\eta=10^{-4}$, it is approximately $7\times$ faster and $1.1$ decades more accurate. These results show that PINS remains efficient and accurate when solving large-scale OT problems at small regularization levels.

\begin{table}[!h]
\caption{$\eta$ sweep on synthetic $n = 400$ for PINS and
Sinkhorn (no EPPA). $\log_{10}$ gap is the final relative gap;
wall is total wall-clock seconds. The rightmost column reports the number of Sinkhorn iterations before either meeting the marginal tolerance or hitting the $50{,}000$-iteration inner cap.}

\label{tab:eta_sweep}
\centering
\small
\begin{tabular}{lrrrrrr}
\toprule
            & \multicolumn{3}{c}{PINS} & \multicolumn{3}{c}{Sinkhorn (no EPPA)} \\
\cmidrule(lr){2-4} \cmidrule(lr){5-7}
$\eta$      & $\log_{10}$ gap & wall (s) & outer iters & $\log_{10}$ gap & wall (s) & iterations \\
\midrule
$10^{-1}$   & $\mathbf{-4.18}$ & $10.9$  & $242$ & $-1.02$ & $0.11$  & ${\sim}100$    \\
$10^{-2}$   & $\mathbf{-4.75}$ & $1.5$   & $60$  & $-2.12$ & $0.10$  & ${\sim}100$    \\
$10^{-3}$   & $\mathbf{-5.39}$ & $\mathbf{3.4}$   & $16$  & $-3.58$ & $267$   & $50{,}000$ (cap)  \\
$10^{-4}$   & $\mathbf{-6.13}$ & $\mathbf{32.2}$  & $4$   & $-5.05$ & $214$   & $50{,}000$ (cap)  \\
\bottomrule
\end{tabular}
\end{table}

\subsection{Large-Scale Benchmarks and Memory Feasibility} \label{sec:numerical_3}
We complement the preceding experiments with two broader comparisons: an accuracy study against Sinkhorn at sizes $n\in\{1024,4096,16{,}384\}$, and a memory-feasibility study against the LP solver \texttt{ot.emd}~\citep{bonneel2011displacement} at sizes up to $n=65{,}536$, where the dense cost matrix alone exceeds 34~GB.

\subsubsection{Benchmark Setup}
We use the DOTmark benchmark~\citep{schrieber2016dotmark} with image sizes $s\in\{32,64,128\}$, so that the corresponding OT problem has dimension $s^2$. For each size, we test three problem classes: CauchyDensity, GRFsmooth, and GRFmoderate. We average over three instance pairs for sizes $32$ and $64$, and use a single canonical instance pair for sizes $128$ and $256$. The reference cost is the squared Euclidean distance on pixel coordinates, normalized so that all entries lie in $[0,1]$.

The memory study uses two cost-matrix families. The DOTmark family, using squared Euclidean costs on a pixel grid, is evaluated at $n\in\{4096,16{,}384\}$. The random family, with entries sampled i.i.d.\ from $\mathrm{Unif}([0,1])$, is evaluated at $n\in\{15{,}000,20{,}000\}$.


\subsubsection{Accuracy Plateau of Sinkhorn-Type Methods}
Sinkhorn returns the optimum of the entropic problem, which differs from the LP optimum by an $\mathcal{O}(\eta\log n)$ regularization bias. At $\eta=10^{-2}$, this bias dominates the inner-iteration error for any practical inner tolerance. Table~\ref{tab:plateau} reports the final log-gap and wall-clock time of each method.

\begin{table}[!h]
\caption{Wall-clock time and final $\log_{10}$ relative error to the LP optimum on the DOTmark benchmark with $\eta = 10^{-2}$. Each row reports one size--class pair. Sinkhorn reaches the entropic plateau in every row, whereas PINS reaches smaller gaps from $-3.77$ to $-5.12$, with harder classes requiring more EPPA outer iterations and reaching higher accuracy.}
\label{tab:plateau}
\centering
\small
\begin{tabular}{llrrrrr}
\toprule
Size & Class & Sk. Time~(s) & Sk. Error~($\log_{10}$) & PINS Time~(s) & PINS Error~($\log_{10}$) \\
\midrule
32  & CauchyDensity & 18.8 & $-2.06$ & 17.7 & $-4.68$ \\
32  & GRFsmooth     & 22.2 & $-2.07$ & 11.6 & $-4.59$ \\
32  & GRFmoderate   & 14.8 & $-2.05$ & 15.8 & $-5.12$ \\
\midrule
64  & CauchyDensity & 232.7 & $-2.05$ &   230.6 & $-3.96$ \\
64  & GRFsmooth     & 335.5 & $-2.05$ &   256.5 & $-4.02$ \\
64  & GRFmoderate   & 382.9 & $-2.04$ &   916.3 & $-4.66$ \\
\midrule
128 & CauchyDensity & 876.1  & $-2.07$ & 2\,236.1 & $-3.84$ \\
128 & GRFsmooth     & 1\,282.6 & $-2.06$ & 1\,632.2 & $-3.77$ \\
128 & GRFmoderate   & 4\,751.1 & $-2.04$ & 8\,157.3 & $-4.14$ \\
\bottomrule
\end{tabular}
\end{table}

\subsubsection{Memory feasibility versus the LP solver}
We next compare the memory feasibility of PINS with the LP solver \texttt{ot.emd}, which uses the POT implementation of network simplex. The peak memory of \texttt{ot.emd} grows roughly as $5n^2$ over the measured range, consisting of the dense cost matrix plus additional network-simplex per-edge state. We measure the physical main memory occupied by the process by resident set size (RSS).

On the four sizes for which both methods were run end-to-end, $n\in\{4096,15{,}000,16{,}384,20{,}000\}$, PINS has $24\%$--$54\%$ lower peak RSS than \texttt{ot.emd} (Table~\ref{tab:memmeas}). Because the random problem lacks exploitable structure, it is slightly more expensive per evaluation than the structured DOTmark cost.

These savings are obtained despite an implementation gap: \texttt{ot.emd} calls a hand-tuned C++ network-simplex routine through POT's compiled extension, whereas PINS is implemented in Python using NumPy and SciPy. The memory advantage is therefore structural rather than engineering.

\begin{table}[!h]
\caption{Peak resident-set memory for PINS versus
\texttt{ot.emd} (POT network simplex).}
\label{tab:memmeas}
\centering
\small
\begin{tabular}{lrrrrr}
\toprule
Problem & $n$ & \texttt{ot.emd} peak RSS (GB) & PINS peak RSS (GB) & PINS / \texttt{ot.emd} \\
\midrule
dotmark\_64    &  4\,096 &   1.38 &   0.63 & 0.46 \\
dotmark\_128   & 16\,384 &  11.22 &   5.29 & 0.47 \\
random\_15000  & 15\,000 &   9.52 &   6.94 & 0.73 \\
random\_20000  & 20\,000 &  16.37 &  12.39 & 0.76 \\
\bottomrule
\end{tabular}
\end{table}

\begin{figure}[!h]
\centering
\includegraphics[width=\columnwidth]{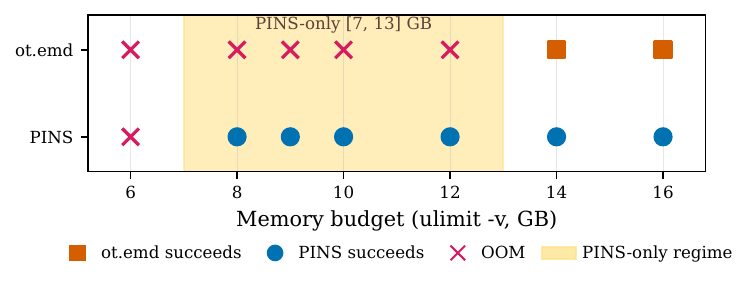}
\caption{Solver feasibility versus per-process memory budget,
at $n = 16{,}384$ (DOTmark size 128). PINS completes for budgets
$\geq 7$~GB; \texttt{ot.emd} completes for budgets $\geq 14$~GB.}
\label{fig:oom_crossover}
\end{figure}

The peak-RSS comparison in Table~\ref{tab:memmeas} gives a direct feasibility statement: for any per-process memory budget between the peaks of the two methods, PINS completes while \texttt{ot.emd} fails. Figure~\ref{fig:oom_crossover} measures this crossover by sweeping the per-process virtual-memory cap, on the $n=16{,}384$ DOTmark instance and re-running both solvers under each cap. PINS completes for any cap at least $7$~GB, whereas \texttt{ot.emd} fails for any cap below $14$~GB. The shaded band marks the $[7,14)$~GB regime in which PINS is the only feasible solver among the two. This is the operational version of the memory advantage: for a machine with a memory budget in this range, the peak-RSS reduction is a hard solvability gap.

\section{Conclusion}
We presented PINS, a two-loop solver for the unregularized OT problem that combines an entropic proximal outer loop with an inner Sinkhorn warm-up and sparse-Newton refinement. The outer loop overcomes the entropic-bias plateau that limits Sinkhorn-type methods, while the inner solver benefits from the quadratic local convergence of Newton's method. The per-iteration Hessian sparsification, which is available at each outer step and independent of the cost matrix, implies cumulative computational savings and makes the method tractable at scale. Across synthetic instances, augmented MNIST, and the DOTmark benchmark, PINS achieves accuracy several orders of magnitude higher than that of Sinkhorn-type baselines at the same $\eta$, is $5$--$73\times$ faster than Sinkhorn under a matched outer loop, and remains memory-feasible on instances where the network-simplex LP solver fails.

\section*{Acknowledgments}
HY was partially supported by the US National Science Foundation under awards DMS-2244988, IIS-2520978, GEO/RISE-5239902, the Office of Naval Research Award N00014-23-1-2007, DOE (ASCR) Award DE-SC0026052, and the DARPA D24AP00325-00. Approved for public release; distribution is unlimited.

\bibliographystyle{plainnat}
\bibliography{references}


\appendix

\section{Limitations}
\label{app:limitations}
The PINS implementation used throughout this paper is single-process Python on top of NumPy and SciPy, with no GPU acceleration. The LP baseline \texttt{ot.emd}, by contrast, is a hand-tuned C++ network-simplex routine accessed through POT's compiled extension; the memory savings reported in Table~\ref{tab:memmeas} are obtained despite this implementation gap. Both the chunked log-sum-exp used for streaming Hessian--vector products and the truncated-CG inner solve are GPU-friendly, so a GPU port and a vectorized batched line search are natural next steps. We expect such an engineering effort to widen, rather than narrow, the gap to the LP solver.

\section{Proof of Theorem~\ref{thm:convergence}}
\label{app:proof_thm1}
The proof relies on the following lemma from \cite{polyak1987introduction}.
\begin{lemma} \label{lem:sequence}
    Let $\{a_k\}_{k=0}^{\infty}$ and $\{b_k\}_{k=0}^{\infty}$ be sequences such that $\{a_k\}$ is bounded from below, $\sum b_k<\infty$, and $a_{k+1}\leq a_k+b_k$ for all $k$. Then $\{a_k\}$ is convergent.
\end{lemma}

\begin{proof}[\textbf{Proof of Theorem \ref{thm:convergence}}]
From (\ref{eq:inexactness}) and the definition of subdifferential, for any $P\in\Omega$ we have
\[\langle \delta^{k+1}-C-\eta(\nabla\phi(X^{k+1})-\nabla\phi(X^k)), P-X^{k+1}\rangle\leq 0\]
Rearranging gives
\[\langle C,X^{k+1}\rangle\leq \langle C,P\rangle+\eta\langle\nabla\phi(X^{k+1})-\nabla\phi(X^k)),P-X^{k+1}\rangle+\langle \delta^{k+1},X^{k+1}-P\rangle.\]
Since $X^{k+1}$ and $P$ are doubly stochastic
\[\langle \delta^{k+1},X^{k+1}-P\rangle \leq \|X^{k+1}-P\|_F\|\delta^{k+1}\|_F\leq\sqrt{2\max\{m,n\}}\epsilon^{k+1}.\]
Moreover, the Bregman identity yields
\[\langle \nabla \phi(X^{k+1})-\nabla \phi(X^k),P-X^{k+1}\rangle=D_{\phi}(P,X^k)-D_{\phi}(P,X^{k+1})-D_{\phi}(X^{k+1},X^k). \]
Combining these, we obtain
\begin{align}\label{eq:descent}
    \langle C,X^{k+1}\rangle\leq \langle C,P\rangle+\eta(D_{\phi}(P,X^k)-D_{\phi}(P,X^{k+1})-D_{\phi}(X^{k+1},X^k))
    +\sqrt{2\max\{m,n\}}\epsilon^{k+1}.
\end{align}
Choosing $P=X^{k}$ in (\ref{eq:descent}) gives
\begin{align*}
    \langle C,X^{k+1}\rangle &\leq \langle C,X^{k}\rangle-\eta(D_{\phi}(X^k,X^{k+1})+D_{\phi}(X^{k+1},X^k))+\sqrt{2\max\{m,n\}}\epsilon^{k+1} \\
    &\leq \langle C,X^{k}\rangle + \sqrt{2\max\{m,n\}}\epsilon^{k+1}.
\end{align*}
Since $\{X^k\}$ is bounded from below, Lemma \ref{lem:sequence} implies convergence.

Next, let $P=X^*$, where $X^*$ is an optimal solution to the original problem (\ref{ot}). From (\ref{eq:descent}),
\begin{align}\label{eq:converge}
    \langle C,X^{k+1}\rangle\leq \langle C,X^*\rangle+\eta(D_{\phi}(X^*,X^k)-D_{\phi}(X^*,X^{k+1})-D_{\phi}(X^{k+1},X^k))
    +\sqrt{2\max\{m,n\}}\epsilon^{k+1}.
\end{align}
The optimality of $X^*$ ensures
\begin{align}
\begin{split}
\label{eq:descent_of_D}
    \eta D_{\phi}(X^*,X^{k+1}) \leq &\eta D_{\phi}(X^*,X^k)-\eta D_{\phi}(X^{k+1},X^k)+\sqrt{2\max\{m,n\}}\epsilon^{k+1} \\
    \leq &\eta D_{\phi}(X^*,X^k)+\sqrt{2\max\{m,n\}}\epsilon^{k+1}.
\end{split}
\end{align}
Thus, $\{D_{\phi}(X^*,X^k)\}$ converges. Taking the limit in (\ref{eq:converge}), and recalling $\epsilon^k \to 0$, the closedness of $\Omega$, and the convergence of $\{X^k\}$, we obtain
\[\lim_{k\to \infty} \langle C,X^{k+1}\rangle\leq \langle C,X^*\rangle.\]
Note that $\Omega$ is bounded, the sequence $\{X^k\} \subset \Omega$ must have at least one convergent subsequence. We denote it as $\{X^{k_j}\}_{j=0}^\infty$, and let its limit point be $\bar{X}$. Since the inner product $\langle C, X \rangle$ is a continuous linear function, the convergence of the subsequence implies
\[\lim_{j\to\infty} \langle C, X^{k_j} \rangle = \langle C, \bar{X} \rangle \leq \langle C,X^*\rangle.\]
Since $\Omega$ is closed, we have $\bar{X}\in \Omega$, and thus $\langle C, \bar{X} \rangle \geq \langle C, X^* \rangle$. Therefore, $\langle C, \bar{X} \rangle = \langle C, X^* \rangle$, which proves that $\bar{X}$ is also an optimal solution to the original OT problem. Substitute $\bar{X}$ for the generic optimal solution $X^*$ in (\ref{eq:descent_of_D}), we establish the entire sequence of Bregman divergences $\{D_{\phi}(\bar{X}, X^k)\}$ is a convergent sequence.  Since $X^{k_j} \to \bar{X}$ and  $0 \log(0) = 0$, we have $\lim_{j\to\infty} D_{\phi}(\bar{X}, X^{k_j}) = 0$. Therefore, $\lim_{k\to\infty} D_{\phi}(\bar{X}, X^k) = 0$. By Pinsker inequality, this shows $\lim_{k\to\infty} \| \bar{X} - X^k \|_1 = 0$. Therefore, the entire matrix sequence $\{X^k\}$ converges exactly to $\bar{X}$, which is an optimal solution of the original OT problem (\ref{ot}). This completes the proof.
\end{proof}

\section{Proof of Theorem~\ref{thm:sparse}}
\label{app:proof_thm2}
\begin{lemma} \label{lem:d1bound} Let $d_1(A,\gB):=\min_{B\in\gB}\|A-B\|_1$. For Problem (\ref{eppa-sub}), if $\eta\leq\frac{\Delta}{R_1+R_{\mathcal{H}}}$, then:
\[d_1(\widetilde{X}^{k*},\gX^{k})\leq2R_1\exp\left(-\frac{\Delta}{\eta R_1}+\frac{R_1+R_{\mathcal{H}}}{R_1}\right).\]
Here $\mathcal{H}(X) = -\sum_{i,j} X_{ij}\log X_{ij}$ is the Shannon entropy, $R_1=\max_{X\in\gP}\|X\|_1$, $R_{\mathcal{H}}=\max_{X,Y\in\gP}\mathcal{H}(X)-\mathcal{H}(Y)$. Specially, we have
$R_1 = 1$, $R_{\mathcal{H}}=\log(mn)$.
\end{lemma}
The proof of this lemma follows from Corollary 9 in \cite{weed2018explicit}. Moreover, since $X$ is a doubly stochastic matrix, it is evident that $\|X\|_1=1$. Furthermore, the maximum value of $\mathcal{H}(X)$ is achieved when $a$ and $b$ are uniformly distributed, and the minimum value is $0$.

\paragraph{Additional notation used in the proof.}
We write
\(
    \gX^{k}:=\argmin_{X\in\mathcal{P}}\inprod{C^k}{X}
\)
for the set of optima of the $k$-th subproblem and
\(
    \widetilde{X}^{k*}:=\argmin_{X\in\mathcal{P}}\,
    \inprod{C^k}{X}+\eta\sum_{ij}X_{ij}\log X_{ij}
\)
for its entropic-regularized optimum. $\widetilde{X}^k$ is the
output of the Sinkhorn phase, and
$X_{\gX}^{k*}\in\argmin_{X\in\gX^k}\|\widetilde{X}^{k*}-X\|_1$.

\begin{proof}[\textbf{Proof of Theorem \ref{thm:sparse}}]
The proof follows from~\citep{tangaccelerating}. We define an approximate Hessian matrix $H^k\in\mathbb{R}^{(m+n)\times(m+n)}$ by:
\begin{align*}
    H^k=-\frac{1}{\eta}\begin{pmatrix}
        \diag(\widetilde{X}^{k} e_n) & X_{\gX}^{k*} \\
        X_{\gX}^{k*}{}^{\top} & \diag(\widetilde{X}^{k}{}^{\top}e_m)
    \end{pmatrix}.
\end{align*}
The number of nonzero entries in $H^k$ is given by $2mn\tau(X_{\gX}^{k*})+m+n$. The difference between $H^k$ and $\nabla^2P^k(\widetilde{X}^k)$ is calculated as:
\begin{align*}
    \|H^k-\nabla^2P^k(\widetilde{X}^k)\|_1=\frac{1}{\eta}\|\widetilde{X}^k-X_{\gX}^{k*}\|_1.
\end{align*}
Using the triangle inequality, the term on the right-hand side can be further bounded:
\begin{align}
    \|\widetilde{X}^{k}-X_{\gX}^{k*}\|_1 &\leq \|\widetilde{X}^{k}-\widetilde{X}^{k*}\|_1+\|X_{\gX}^{k*}-\widetilde{X}^{k*}\|_1 
    \leq \|\widetilde{X}^{k}-\widetilde{X}^{k*}\|_1+d_1(\widetilde{X}^{k*},\gX^{k}). \label{eq:bounddcp}
\end{align}
The first term in (\ref{eq:bounddcp}) can be bounded by the Pinsker inequality and Theorem 4.4 in \cite{ghosal2022convergence}. There exists some constants $\kappa,l$, such that for any Sinkhorn steps exceeding $\kappa$, 
$\|\widetilde{X}^{k}-\widetilde{X}^{k*}\|_1\leq\frac{\sqrt{l}}{\sqrt{T_1}}.$ The second term in (\ref{eq:bounddcp}) could be bounded by Lemma \ref{lem:d1bound} when $\eta\leq\frac{\Delta}{1+\log(mn)}$:
\begin{align*}
    d_1(\widetilde{X}^{k*},\gX^{k})&\leq 2\exp(-\frac{\Delta}{\eta}+1+\log(mn)) \leq 6mn\exp(-\frac{\Delta}{\eta}).
\end{align*}

Thus, combining the bounds, the Hessian matrix approximation follows directly.

Finally, based on Theorem 3.7 in \cite{dieci2019boundary}, there is a unique solution to the original unregularized problem almost surely (with probability one). If $m=n$, the feasible set defines a Birkhoff-von Neumann polytope, spanning an $(n-1)^2$-dimensional affine subspace of $\sR^{n^2}$, whose vertices are precisely the permutation matrices \citep{stanley2011enumerative}. According to Theorem 2.7 in \cite{bertsimas1997introduction}, the linear objective's minimum over a nonempty polyhedron is achieved at an extreme point. Thus $\tau(X_{\gX}^{k*})\leq\frac{n^2-(n-1)^2}{n^2}=\frac{2n-1}{n^2}$, which then implies $\tau(H^k)\leq\frac{2n+2(2n-1)}{(2n)^2}=\frac{3n-1}{2n^2}$.
\end{proof}

\section{Color Transfer Experiment}
\label{app:color_transfer}

OT is widely used in imaging~\citep{rubner2000earth,pitie2007automated,merigot2011multiscale,bonneel2013example,
solomon2015convolutional,ferradans2013static,ferradans2014regularized,bonneel2015sliced,
papadakis2015optimal,bao2024optimal,zhang2025hot}, where small transport errors map to visible artifacts. We illustrate sub-plateau accuracy on the color-transfer
task~\citep{pitie2007automated}, a setting where small transport
errors translate into visible artifacts of the transferred image.

\paragraph{Setup.}
Following~\citet{pmlr-v84-blondel18a}, we quantize the source and target images into $m=n=256$ RGB-centroid clusters. The marginals $a$ and $b$ are the normalized cluster sizes, and the cost is the squared Euclidean distance between centroids. After solving the OT problem, source colors are remapped using the standard barycentric projection~\citep{pitie2007automated}. We compare PINS with Sinkhorn at $\eta=1$ on three image pairs (\textit{autumn}, \textit{grafiti}, and \textit{comunion}). We report content fidelity using SSIM~\citep{wang2004image} and style fidelity using VGG style loss~\citep{gatys2016image}. All other hyperparameters match those in \S\ref{sec:numerical_1}.

\begin{figure}[!h]
  \centering
  \includegraphics[width=.32\linewidth]{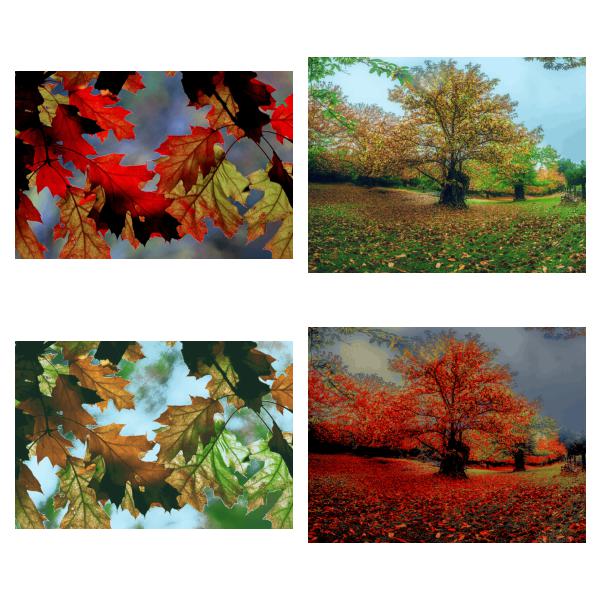}\hfill
  \includegraphics[width=.32\linewidth]{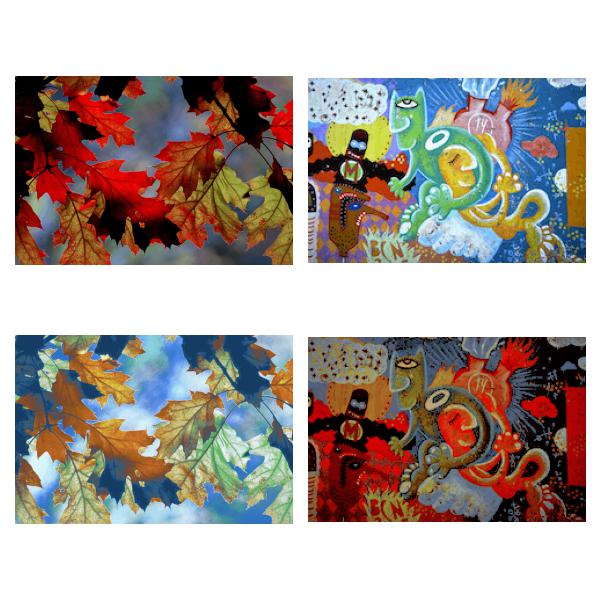}\hfill
  \includegraphics[width=.32\linewidth]{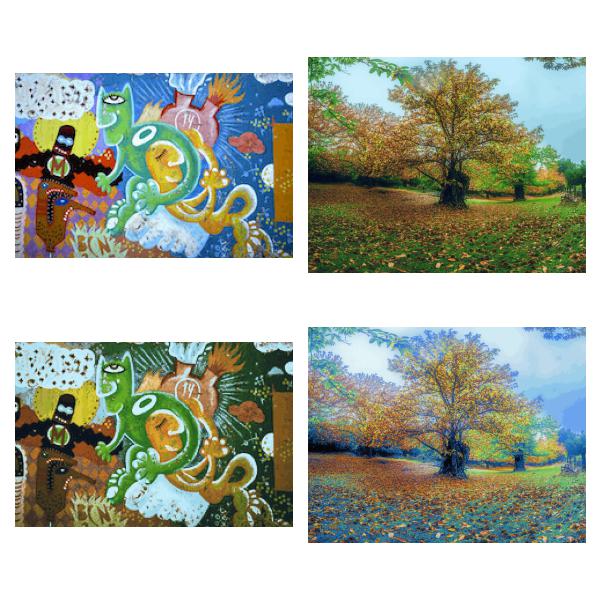}
  \caption{Bidirectional color transfer with PINS. Top row: original source and target images; bottom row: transferred outputs. Each column corresponds to one source--target pair.}
  \label{fig:color_transfer}
\end{figure}

\paragraph{Results.}
Across all three image pairs, PINS improves SSIM by $0.16$--$0.31$ over Sinkhorn and reduces VGG style loss by up to $38\%$ (Table~\ref{tab:color_transfer}). The style-loss margin is smaller on the Grafiti--Comunion pair, but the SSIM improvement of PINS is uniform and consistent in both transfer directions. This suggests that the gain comes from a more accurately resolved transport plan rather than from one direction dominating the average.

\begin{table}[!h]
\centering
\caption{Color transfer with PINS versus Sinkhorn at $\eta=1$. Content similarity is measured by SSIM, where higher is better; style similarity is measured by VGG style loss, where lower is better. Each value averages the two transfer directions for the pair. The best value for each metric is in bold.}
\label{tab:color_transfer}
\begin{tabular}{l|cc|cc}
\hline
\multirow{2}{*}{Image pair} & \multicolumn{2}{c|}{SSIM $\uparrow$}
                            & \multicolumn{2}{c}{Style loss $\downarrow$} \\
                            & PINS & Sinkhorn & PINS & Sinkhorn \\
\hline
Autumn--Grafiti     & $\mathbf{0.53}$ & $0.29$ & $\mathbf{0.18}$ & $0.29$ \\
Autumn--Comunion    & $\mathbf{0.54}$ & $0.37$ & $\mathbf{0.14}$ & $0.17$ \\
Grafiti--Comunion   & $\mathbf{0.76}$ & $0.45$ & $\mathbf{0.34}$ & $0.35$ \\
\hline
\end{tabular}
\end{table}

\section{Numerical Details and Additional Results}
\label{app:section_5_1_extra_figures}

\subsection{Implementation Details for \S\ref{sec:numerical_1}}
\label{sec:numerical_1_implementation}

\paragraph{Hyperparameters.} On synthetic instances, we use $\eta = 10^{-2}$ unless stated otherwise; on augmented MNIST, we use $\eta = 10^{-1}$. The Sinkhorn warm-up uses $T_1 = 100$ inner block updates, and the sparse-Newton inner solver uses $T_2 = 50$ damped Newton steps with Armijo backtracking ($\beta = 0.5$, $c_1 = 10^{-4}$). The EPPA outer cap is $K = 250$. The sparsification threshold $\rho$ is chosen so that the retained fraction matches the bound in Theorem~\ref{thm:sparse}, namely at most $(3n-1)/(2n^2)$ entries of the block. Truncated CG is run to a relative residual of $10^{-6}$. The marginal stopping tolerance is $10^{-9}$.

\paragraph{Reference solution.} The relative gap reported in Section \ref{sec:numerical_1} is computed against the LP optimum returned by \texttt{ot.emd} from POT~\citep{bonneel2011displacement}, which solves the unregularized problem~(\ref{ot}) by network simplex.

\paragraph{Sinkhorn iteration cap.} The plain log-domain Sinkhorn baseline runs to a marginal tolerance of $10^{-9}$ or to a hard cap of $50{,}000$ iterations, whichever occurs first. The cap is reached in the small-$\eta$ rows of Table~\ref{tab:eta_sweep}, reflecting the standard $\Theta(1/\eta)$ slowdown of Sinkhorn rather than a numerical failure.

\paragraph{Hardware.} All Section \ref{sec:numerical_1} runs were
performed single-threaded on an Intel Xeon Gold 6248 node at 2.5 GHz with 384 GB RAM, using NumPy 1.26 on top of MKL. The DOTmark
memory-feasibility study in Section \ref{sec:numerical_3} reports peak RSS measured by \texttt{resource.getrusage} after the solver returns.

\subsection{Additional Results}

\begin{figure}[!ht]
\centering
\includegraphics[width=0.8\columnwidth]{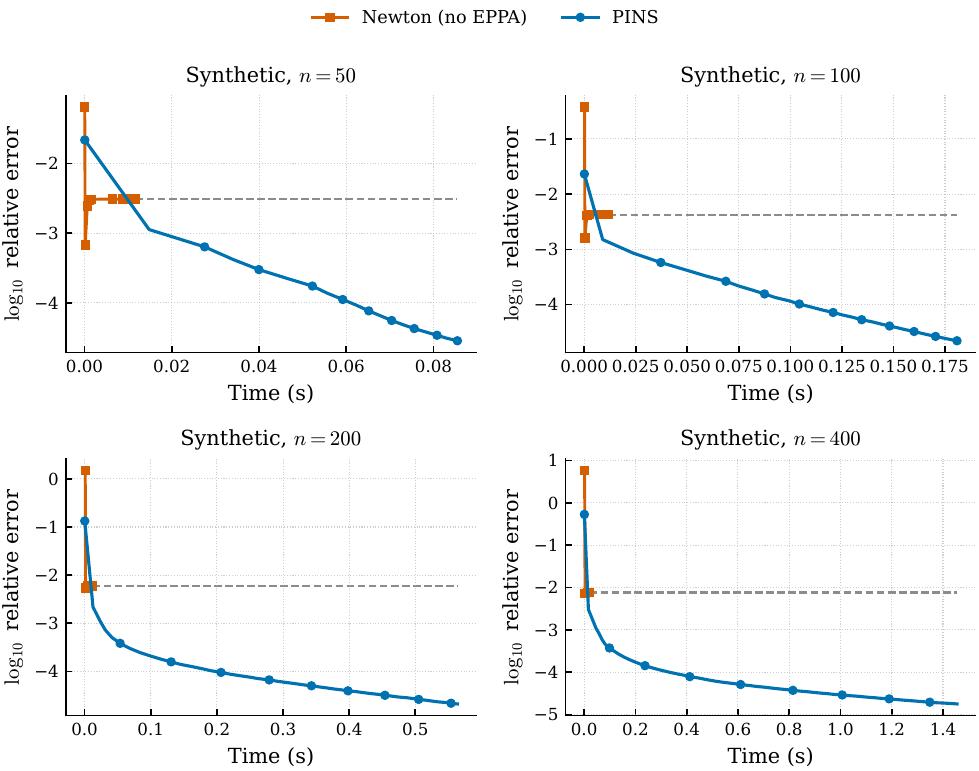}
\caption{Per-iteration $\log_{10}$ relative error to the LP optimum on synthetic data with $\eta = 10^{-2}$; panels correspond to $n \in \{50, 100, 200, 400\}$. Each curve plots the per-iteration log-gap of one method against cumulative wall-clock time. A line break on the PINS curve corresponds to one EPPA outer iteration; a line break on the Newton-without-EPPA curve corresponds to one inner Newton iteration. Markers are subsampled for readability and not one-to-one with iterations. The dashed gray segment on the Newton curve is a constant extension at its final log-gap, indicating that the no-EPPA baseline cannot improve further with additional wall-clock budget. Numerical values are reported in Table~\ref{tab:final_accuracy}.}

\label{fig:eppa_synthetic_appx}
\end{figure}

\begin{figure}[!ht]
\centering
\includegraphics[width=0.8\columnwidth]{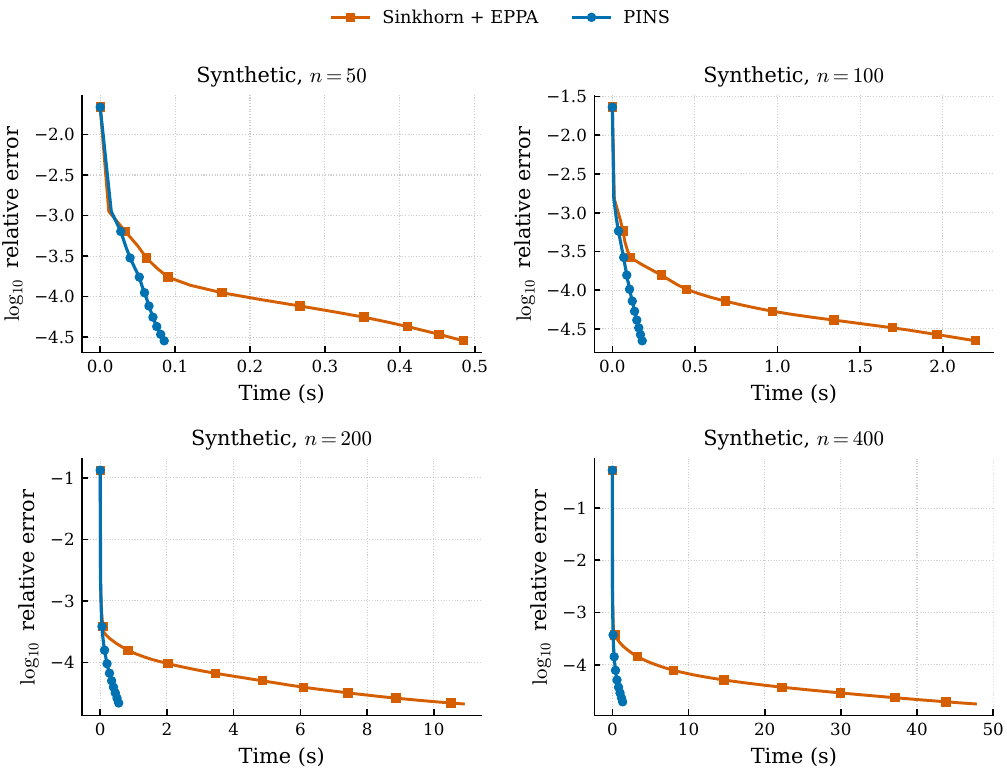}
\caption{PINS versus Sinkhorn + EPPA on synthetic data with $\eta = 10^{-2}$; panels correspond to $n \in \{50, 100, 200, 400\}$. Each line break on the Sinkhorn + EPPA curve corresponds to one EPPA outer iteration of the Sinkhorn-inner variant. Numerical values are reported in Table~\ref{tab:speedups}.}
\label{fig:efficiency_synthetic_appx}
\end{figure}



\end{document}